\pdfoutput=1

\documentclass[11pt]{article}

\usepackage[preprint]{acl}

\usepackage{times}
\usepackage{latexsym}

\usepackage[T1]{fontenc}

\usepackage[utf8]{inputenc}

\usepackage{microtype}

\usepackage{inconsolata}

\usepackage{graphicx}

\usepackage{listings}
\usepackage{xcolor}

\usepackage{footmisc}

\definecolor{codegreen}{rgb}{0,0.6,0}
\definecolor{codegray}{rgb}{0.5,0.5,0.5}
\definecolor{codepurple}{rgb}{0.58,0,0.82}
\definecolor{backcolour}{rgb}{0.95,0.95,0.92}

\lstdefinestyle{mystyle}{
    backgroundcolor=\color{backcolour},   
    commentstyle=\color{codegreen},
    keywordstyle=\color{magenta},
    numberstyle=\tiny\color{codegray},
    stringstyle=\color{codepurple},
    basicstyle=\ttfamily\footnotesize,
    breakatwhitespace=false,         
    breaklines=true,                 
    captionpos=b,                    
    keepspaces=true,                 
    numbers=left,                    
    numbersep=5pt,                  
    showspaces=false,                
    showstringspaces=false,
    showtabs=false,                  
    tabsize=2
}

\title{PyFCG: Fluid Construction Grammar in Python}

\author{Paul Van Eecke\thanks{Both authors contributed equally. The authors declare that this paper was conceived and written without the assistance of generative writing aids.}\\
  Artificial Intelligence Laboratory \\
  Vrije Universiteit Brussel, Belgium \\
  \texttt{paul@ai.vub.ac.be} \\\And
  Katrien Beuls\footnotemark[1] \\
  Faculté d'informatique \\
  Université de Namur, Belgium \\
  \texttt{katrien.beuls@unamur.be} \\}

\begin{document}
\maketitle

\begin{abstract}
We present PyFCG, an open source software library that ports Fluid Construction Grammar (FCG) to the Python programming language. PyFCG enables its users to seamlessly integrate FCG functionality into Python programs, and to use FCG in combination with other libraries within Python's rich ecosystem. Apart from a general description of the library, this paper provides three walkthrough tutorials that demonstrate example usage of PyFCG in typical use cases of FCG: 
(i) formalising and testing construction grammar analyses, (ii) learning usage-based construction grammars from corpora, and (iii) implementing agent-based experiments on emergent communication.
\end{abstract}

\section{Fluid Construction Grammar}

Fluid Construction Grammar \citep[FCG --][]{steels2004constructivist,vantrijp2022fcg,beuls2023fluid} is a computational construction grammar framework that provides a collection of high-level building blocks for representing, processing and learning fully-operational construction grammars.  The FCG framework is conceived as an open instrument that is not tied to a particular construction grammar theory, but that strives for compatibility with any linguistic theory that adheres to the most fundamental tenets underlying constructionist approaches to language \citep[see e.g.][]{fillmore1988mechanisms,croft2001radical,goldberg2003constructions}. As such, it subscribes to the view (i) that language users dynamically build up their own linguistic systems as they communicate with other members of their community,  (ii) that these linguistic systems can be captured as a network of form-meaning mappings called constructions, and (iii) that these constructions can pair forms and meanings of arbitrary complexity and degree of abstraction, thereby facilitating a uniform handling of both compositional and non-compositional linguistic phenomena. 

FCG is primarily being used as the language representation, processing and learning component in agent-based models of linguistic communication.  Such models simulate the emergence, evolution and acquisition of human languages in populations of artificial agents that take part in situated communicative interactions modelled after those that human language users continuously engage in \citep[e.g.][]{vantrijp2016evolution,beuls2024humans}. Other common uses of FCG include the formalisation and computational operationalisation of construction grammar analyses \citep[e.g.][]{gerasymova2012expressing,micelli2012field}, and the corroboration of construction grammar theories with empirical data \citep[e.g.][]{moerman2024evaluating}. 

FCG is being developed as an open-source community project, which brings together the construction grammar and computational linguistics communities. While strong ties between both communities already existed when the field of construction grammar was founded in the 1980s, recent initiatives such as UCxn \citep{weissweiler2024ucxn} and the Construction Grammars and NLP (CxGs+NLP) workshop series \citep{bonial2023proceedings}, along with an increasing volume of work on constructions in Large Language Models \citep[see e.g.][]{madabushi2020cxgbert,tseng2022cxlm,weissweiler2022better,bonial2024constructing,xu2024coelm}, bear witness to a growing interest in research at the intersection of both fields.

\section{FCG in Python, Really?}

Readers who have regularly used FCG might argue that there already exists a stable and mature, efficient, cross-platform and open source implementation of FCG, with an active and dedicated, albeit small, developer community\footnote{See \url{https://gitlab.ai.vub.ac.be/ehai/babel} for the project's code repository.}. Indeed, the reference FCG implementation is written in Common Lisp, a dynamic and extensible programming language that, admittedly, excellently fits the project's requirements, including multi-paradigm, high-level and multi-threaded programming, fast prototyping, and highly efficient symbol processing. So why would anyone spend time and effort on a Python port?  

The reality is that Python has become the most popular programming language in the world\footnote{\url{https://www.tiobe.com/tiobe-index/}} and that, more consequentially, it has also become the dominant language in programming education, as well as today's de facto standard in both linguistics and natural language processing research. Practically speaking, this entails that the success of a software library targeted at the broader computational linguistics community depends before all other things on its compatibility with the Python ecosystem. While this might sound utterly unreasonable, it is not entirely so. For one thing, computer programs typically integrate a variety of external libraries, and, for better or for worse, the largest selection of libraries tends to be developed for the most popular programming languages. For another, the investment involved in learning to use a new programming language and environment is a very real obstacle for potential users in today's time-pressed society. 

The development and release of PyFCG follows a trend set by many other libraries that are commonly used in the computational linguistics community. For example, the Stanford CoreNLP Java library \citep{manning2014stanford} is now accessible from Python through the Stanza library \citep{qi2020stanza}. Likewise, the PRAAT system for phonetic analysis \citep{boersma2025praat}, written in C and C++, is now widely used in Python programs via the Parselmouth library \citep{jadoul2018introducing}. The Torch library for tensor computation, written in C and Lua, is now even primarily being developed for Python as part of the PyTorch project \citep{paszke2019pytorch}.

\section{FCG in Python, Finally!}

Readers who have not yet used FCG, perhaps for the reasons mentioned above, might wonder why it has taken so long for FCG to make its way into the Python ecosystem. It could be ascribed to a lack of actual problems with the existing reference implementation, to the sheer size, scope and complexity of its codebase, to the difficulty of funding scientific software development, or perhaps most likely, to a combination of all these factors. But fret no more:

\begin{lstlisting}[language=bash]
$ pip install pyfcg
\end{lstlisting}

Once pip-installed, PyFCG can readily be imported as a module into Python programs. It is customary to define \texttt{fcg} as an alias for the PyFCG module, so that all functionality is available within the \texttt{fcg} namespace. We initialise PyFCG by calling its \texttt{init()} function, which loads (or downloads if necessary) all external dependencies. PyFCG's documentation is available on the \textit{Read the Docs} platform\footnote{\url{https://pyfcg.readthedocs.io}} and interactive tutorial notebooks supporting this paper can be downloaded from the FCG community website\footnote{\url{https://fcg-net.org/pyfcg}\label{fn:fcg}}.

\begin{lstlisting}[language=Python]
>>> import pyfcg as fcg
>>> fcg.init()
\end{lstlisting}

On the highest level, PyFCG defines three classes that are of interest to the user: \texttt{Agent}, \texttt{Grammar} and \texttt{Construction}. The idea is that an agent (of type \texttt{Agent}) has a grammar (of type \texttt{Grammar}), which in turn holds constructions (of type \texttt{Construction}). The \texttt{Agent} class is the main entry point for the user. Upon the creation of a new agent, it is automatically initialised with an empty grammar, i.e. a grammar that holds zero constructions.

\begin{lstlisting}[language=Python]
>>> demo_agent = fcg.Agent()
>>> demo_agent.grammar.size()
0
\end{lstlisting}

It was an explicit design choice to tie grammars to agents, emphasising FCG's view that a grammar always represents the linguistic knowledge of an individual language user. Grammars are in principle never shared between agents and cannot exist outside an agent. Instances of the \texttt{Grammar} class should therefore only be created implicitly via the \texttt{Agent} class. 

\section{PyFCG at Work}

We present three walkthrough tutorials that showcase how PyFCG can be integrated in typical use cases of FCG: grammar formalisation and testing (\ref{sec:grammar-writing}), learning grammars from semantically annotated corpora (\ref{sec:corpora}), and modelling emergent communication (\ref{sec:experiments}). Each tutorial is accompanied by an interactive notebook, which can be downloaded from the FCG community website\footref{fn:fcg}.

\subsection{Grammar formalisation and testing}
\label{sec:grammar-writing}

A common use of FCG revolves around the formalisation and computational operationalisation of construction grammar theories and analyses. Not only can computational operationalisations help validate their preciseness and internal consistency, they also facilitate the comparison, exchange and integration of insights from different researchers \citep{vantrijp2022fcg}. This tutorial exemplifies how PyFCG can be used to equip an agent with a designed grammar, how new constructions can be added to or removed from the agent's grammar on the fly, how the agent can use its grammar to comprehend and formulate utterances, and how all these processes can be visually inspected through FCG's graphical web interface. For more information on aspects of the syntax and semantics of FCG that are not particular to the PyFCG module, we refer the interested reader to \citet[][Chapter 3]{vaneecke2018generalisation}\footnote{Or alternatively to \url{https://emergent-languages.org/wiki/docs/recipes/fcg/syntax-and-semantics}.}.

After importing and initialising PyFCG, we create a new agent named Sue. Sue, as an instance of the \texttt{fcg.Agent} class, is automatically initialised with an empty grammar. Sue is also assigned a unique identifier:

\begin{lstlisting}[language=Python]
>>> sue = fcg.Agent(name='Sue')
>>> sue
<Agent: Sue (id: sue-1) ~ 0 cxns>
\end{lstlisting}

Sue can read in a predefined grammar, specified in the Open FCG Exchange Format (OFEF). In this case, we use PyFCG's \texttt{load\_resource} function to download a human-designed demo grammar fragment that specifies six constructions for processing the English resultative sentence ``\textit{Firefighters cut the child free}.''\footnote{Example after \citet{hoffmann2018creativity}.}. This grammar fragment  uses the Abstract Meaning Representation format \citep[AMR;][]{banarescu2013abstract} to represent constructional meaning\footnote{AMR meaning representation kindly provided by Claire Bonial (p.c. 31/03/2023).}. We instruct Sue to load the grammar specified in the file by calling their \texttt{load\_grammar\_from\_file} method and then list the names of the constructions that were loaded.

\begin{lstlisting}[language=Python]
>>> f = fcg.load_resource('demo-resultative.json')
>>> sue.load_grammar_from_file(f)
>>> sue
<Agent: Sue (id: sue-1) ~ 6 cxns>
>>> list(sue.grammar.cxns.keys())
['firefighters-cxn', 'child-cxn', 
 'cut-cxn', 'free-cxn', 'np-cxn', 
 'resultative-cxn']
\end{lstlisting}

\begin{figure*}
\includegraphics[width=\textwidth]{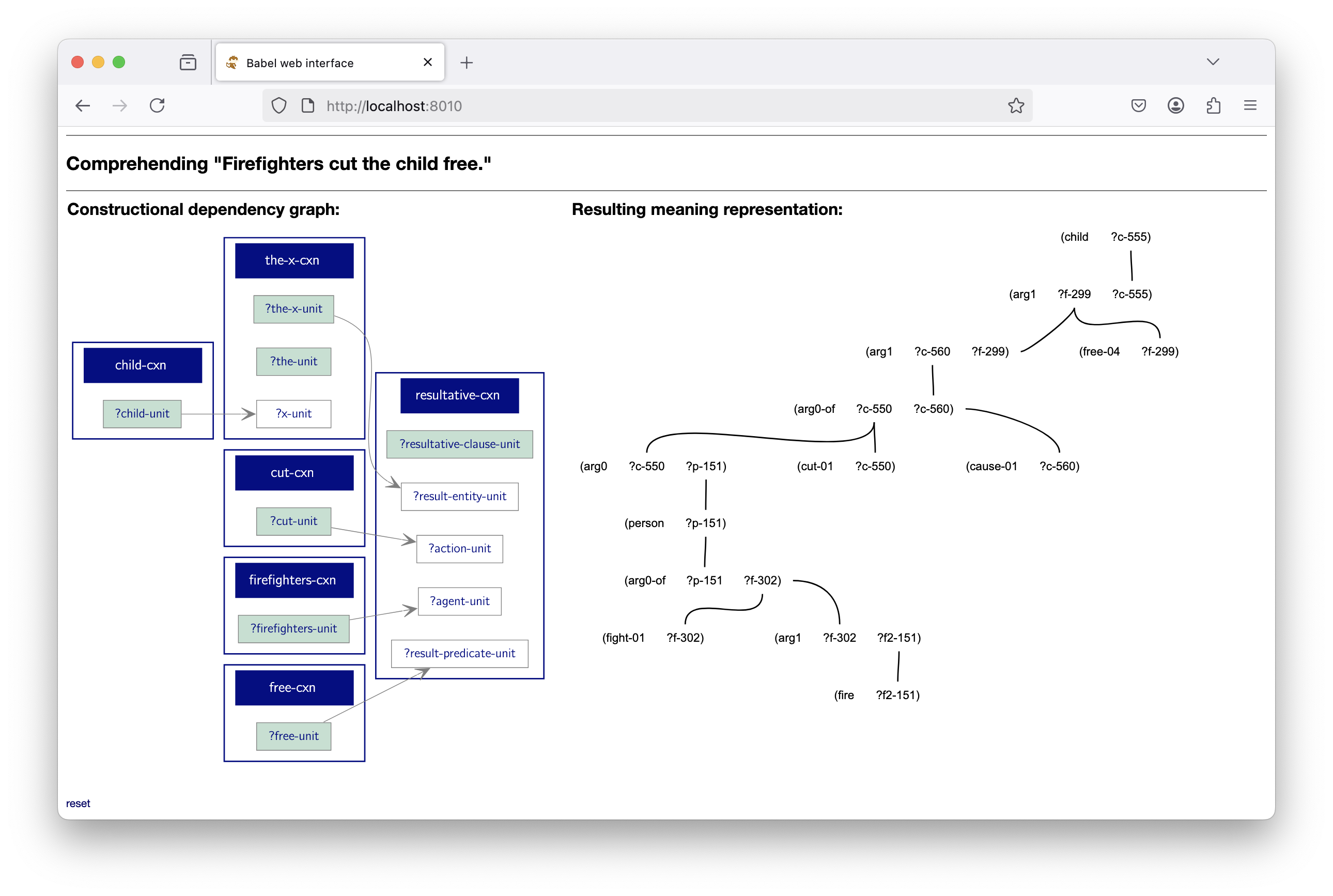}
\caption{Screen capture of FCG's web interface after calling  \texttt{sue.comprehend("Firefighters cut the child free.")} with the \texttt{trace-fcg} monitor activated.}
\label{fig:wi}
\end{figure*}

We now ask Sue to comprehend the utterance ``\textit{Firefighters cut the child free}.''. In FCG, comprehension and formulation respectively refer to the processes of mapping utterances to their meaning representation and vice versa. In order to be able to visually inspect Sue's comprehension process, we first start up FCG's graphical web interface and activate the standard \texttt{trace-fcg} monitor. After having comprehended the utterance, we visualise the resulting AMR meaning representation in the more human-readable Penman format. We can see that Sue understood that a cutting action in the sense of `\textit{slice, injure}` \citep[denoted by PropBank's \texttt{cut-01} roleset;][]{palmer2005proposition} was performed by an `\textit{intentional cutter}' (\texttt{arg0} in \texttt{cut-01}), more in particular a person who habitually `\textit{fights}' (\texttt{fight-01}) fire (\texttt{arg1} in \texttt{fight-01}), and that the cutting action itself led to (\texttt{arg0-of} in \texttt{cause-01})  the `\textit{unconstrained, unrestricted}' state (\texttt{free-04}) of a child (\texttt{arg1} in \texttt{free-04}).  A screen capture of the web interface after comprehending the utterance is shown in Figure \ref{fig:wi}.

\begin{lstlisting}[language=Python]
>>> fcg.start_web_interface()
>>> fcg.activate_monitors(['trace-fcg'])
>>> amr = sue.comprehend("Firefighters cut the child free.")
>>> fcg.predicate_network_to_penman(amr)
'(c / cut-01
      :arg0 (p / person
           :arg0-of (f / fight-01
                :arg1 (f2 / fire)))
      :arg0-of (c2 / cause-01
           :arg1 (f3 / free-04
                :arg1 (c3 / child))))'
\end{lstlisting}

Let us now add a new construction to Sue's grammar: the \textsc{dog-cxn} that in essence pairs the form ``\textit{dog}'' with its AMR meaning of instantiating the \texttt{dog} concept. Along with its name, we specify its contributing and conditional poles, and load it into Sue. We also add a categorial link between the category proper to the \textsc{dog-cxn} and the category of the noun slot in the \textsc{np-cxn}, following the design choices made in the demo grammar fragment and very much in the spirit of Radical Construction Grammar \citep{croft2001radical}.

\begin{lstlisting}[language=Python]
>>> dog_cxn = fcg.Construction(
      name= 'dog-cxn', 
      contributing_pole= 
        [('?dog-unit',
          {'referent': '?d',
           'category': 'dog-cxn',
           'boundaries': 
                 ('?left', '?right')})],
      conditional_pole= 
        [('?dog-unit',
          {'#meaning': [('dog', '?d')]},
          {'#form': 
             [('sequence', '"dog"', 
               '?left', '?right')]})])
>>> sue.add_cxn(dog_cxn)
>>> sue.add_category('dog-cxn')
>>> sue.add_link('dog-cxn','np-cxn-n')
>>> sue
<Agent: Sue (id: sue-1) ~ 7 cxns>
\end{lstlisting}

We now ask Sue to use their grammar to \texttt{formulate} an utterance that expresses that a cutting action in the sense of `\textit{slice, injure}` was performed by an `\textit{intentional cutter}', more in particular a person who habitually `\textit{fights}' fire, and that the cutting action itself led to the `\textit{unconstrained, unrestricted}' state of a dog:

\begin{lstlisting}[language=Python]
>>> amr = '(c / cut-01
             :arg0 (p / person
                 :arg0-of (f / fight-01
                     :arg1 (f2 / fire)))
             :arg0-of (c2 / cause-01
                 :arg1 (f3 / free-04
                     :arg1 (d / dog))))'
>>> amr = fcg.penman_to_predicate_network(amr)
>>> sue.formulate(amr)
'Firefighters cut the dog free.' , 
\end{lstlisting}

We can see that Sue produces the utterance ``\textit{Firefighters cut the dog free.}''. Again, the construction application process can be traced in detail in the web interface. 

\subsection{Learning grammars from corpora}
\label{sec:corpora}

A second use case of FCG concerns the learning of construction grammars from corpora of language use. We take the example of \texttt{fcg-propbank}, an existing FCG subsystem for learning construction grammars from PropBank-annotated corpora. We demonstrate how a pretrained grammar comprising tens of thousands of constructions can be loaded into an agent and used to extract semantic frames from open-domain text, how a new grammar can be learnt from annotated data, and how large grammars can be saved in an efficiently loadable binary format.

As always, we start by creating an agent. In this case, the agent is an instance of the \texttt{fcg.PropBankAgent} class, a subclass of the \texttt{fcg.Agent} class provided by the \texttt{fcg-propbank} subsystem. We download a pretrained, precompiled grammar for English and load it into our agent using its \texttt{load\_grammar\_image} method. The agent now has at its disposal a grammar consisting of 21,052 constructions.

\begin{lstlisting}[language=Python]
>>> pb_pretrained = fcg.PropBankAgent()
>>> f = fcg.load_resource('pb-en.store')
>>> pb_pretrained.load_grammar_image(f)
>>> pb_pretrained
<Agent: (id: agent-1) ~ 21052 cxns>
\end{lstlisting}

Our agent can now use its pretrained grammar to comprehend new utterances. Below, we instruct our agent to comprehend the passive utterance ``\textit{Margaret Thatcher was elected Prime Minister of Britain.}'' \citep[][from \textsc{now}-19-12-08-\textsc{us}]{herbst2024construction}. The resulting meaning representation reveals that the agent identified a single semantic frame  that instantiates the \texttt{elect.01} PropBank roleset (`\textit{elect someone to an office or position}'). The agent also understood that the roles of `\textit{candidate}' (\texttt{arg1}) and `\textit{office or position}' (\texttt{arg2}) in this instance of \texttt{elect.01} are respectively taken up by ``\textit{Margaret Thatcher}'' and ``\textit{Prime Minister of Britain}''.

\begin{lstlisting}[language=Python]
>>> pb_pretrained.comprehend("Margaret Thatcher was elected Prime Minister of Britain.")
[{'roleset': 'elect.01', 'roles': [
  ('v', "elected"),
  ('arg1', "Margaret Thatcher"),
  ('arg2', "Prime Minister of Britain")]
 }]
\end{lstlisting}

\begin{figure}
\centering
\includegraphics[width=.9\columnwidth]{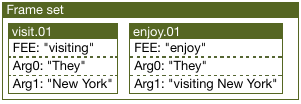}
\caption{Semantic frames resulting from the comprehension process of the utterance ``\textit{They enjoy visiting New York.}'' (COCA).}
\label{fig:frames}
\end{figure}

To enhance human readability, we can again choose to activate an FCG monitor to trace the comprehension process in the web interface. Figure \ref{fig:frames} shows the visualisation of the two frames extracted from the utterance ``\textit{They enjoy visiting New York}'' (COCA).

Let us now create a second agent, again as an instance of the \texttt{fcg.PropBankAgent} class, but let it learn a new grammar from corpus data instead of loading a pretrained one. After having downloaded an example CoNNL file, in which a number of English sentences are annotated with PropBank rolesets\footnote{Due to licensing restrictions, we are not able to provide large PropBank-annotated corpora as downloadable PyFCG resources. We invite interested readers to obtain such corpora (e.g. OntoNotes or EWT) directly from the \textit{Linguistic Data Consortium}.}, we call the agent's \texttt{learn\_grammar\_from\_conll\_file} method. This call initiates the learning process implemented by the \texttt{fcg-propbank} subsystem and equips the agent with the resulting grammar. In this case, the agent has learnt two lexical constructions (for verbs with the lemmas \textit{give} and \textit{send}), two word sense constructions (for the rolesets \texttt{give.01} and \texttt{send.01}), and two argument structure constructions (a double object construction and a prepositional dative construction). 

\begin{lstlisting}[language=Python]
>>> pb_learner = fcg.PropBankAgent()
>>> f = fcg.load_resource('pb-annotations.conll')
>>> pb_learner.learn_grammar_from_conll_file(f)
>>> pb_learner
<Agent: (id: agent-2) ~ 6 cxns>
>>> list(pb_learner.grammar.cxns.keys())
['give(v)-cxn', 'send(v)-cxn',
 'give.01-cxn', 'send.01-cxn',
 'arg0(np)+v(v)+arg2(np)+arg1(np)-cxn',
 'arg0(np)+v(v)+arg1(np)+arg2(pp)-cxn']
\end{lstlisting}

We can now instruct our agent to comprehend a previously unseen utterance, using the grammar it just learnt, by calling its \texttt{comprehend} method. While comprehending ``\textit{The King of the Belgians sent a box of chocolates to Forrest Gump.}'', the agent identifies an instance of the \texttt{send.01} (`\textit{give}') roleset, with ``\textit{The King of the Belgians}'' as the `\textit{sender}' (\texttt{arg0}), ``\textit{a box of chocolates}'' as the `\textit{thing sent}' (\texttt{arg1}) and ``\textit{to Forrest Gump}'' as the `\textit{sent-to}' entity (\texttt{arg2}). 

\begin{lstlisting}[language=Python]
>>> pb_learner.comprehend("The King of the Belgians sent a box of chocolates to Forrest Gump.")
[{'roleset': 'send.01',
  'roles': [('v', "sent"),
   ('arg0', "The King of the Belgians"),
   ('arg1', "a box of chocolates"),
   ('arg2', "to Forrest Gump")]}]
\end{lstlisting}

After learning a grammar, it can be saved by calling the \texttt{save\_grammar\_image} method of the \texttt{fcg.Agent} class. This method saves the grammar to a file in a compiled, binary format that can later efficiently be loaded using an agent's \texttt{load\_grammar\_image} method.

\begin{lstlisting}[language=Python]
>>> propbank_agent.save_grammar_image('usage-based-grammar.store')
>>> new_agent = fcg.PropBankAgent()
>>> new_agent.load_grammar_image('usage-based-grammar.store')
<Agent: (id: agent-3) ~ 6 cxns>
\end{lstlisting}

\subsection{Modelling emergent communication}
\label{sec:experiments}

The final walkthrough tutorial exemplifies the primary use case of FCG: implementing the linguistic capability of autonomous agents in agent-based models of emergent communication. In such experiments, agents start out with an empty grammar and gradually build up their linguistic knowledge as they take part in situated communicative interactions with other agents in the population. This tutorial presents a PyFCG-powered implementation of the canonical naming game experiment \citep{steels1996perceptually,vaneecke2022language}, in which a population of agents converges on a naming convention used to refer to objects in their environment. The choice for the naming game was made for didactic reasons, as implementations of language games involving the emergence of more complex grammars, where the use of a framework like FCG truly comes to its own, soon become strenuous to read through.

A first step in setting up a language game experiment concerns the creation of a population of agents. We define our agents as instances of a new class \texttt{NGAgent} that subclasses from PyFCG's \texttt{fcg.Agent} class. The agents are thereby initialised with an empty grammar and inherit a collection of methods for interacting with instances of the \texttt{fcg.Grammar} and \texttt{fcg.Construction} classes. 

\begin{lstlisting}[language=Python]
class NGAgent(fcg.Agent):
  ...
>>> NGAgent()
<Agent: agent (id: agent-1) ~ 0 cxns>
\end{lstlisting}

We also define a new experiment class \texttt{NGExperiment}. Upon initialisation, a population is created as a set of \texttt{NGAgent} instances,  and a world is created as a set of abstract objects. Two methods are also associated to this class. The \texttt{run\_interaction} method (cf. below) initiates a new communicative interaction as an instance of the \texttt{NGInteraction} class, makes the interaction happen, and records its outcome. The \texttt{run\_series} method runs a given number of interactions.

\begin{lstlisting}[language=Python]
class NGExperiment():
 def __init__(self, configuration={}):
  ...
  self.world = 
  ['obj-%d' % i for i in range(configuration['nr_of_objects'])]
  self.population = 
  [NGAgent() for i in range(configuration['nr_of_agents'])]
		
 def run_interaction(self):
  ci = NGInteraction(self)
  ci.interact()
  ci.record_communicative_success()
  ci.record_lexicon_size()
  ci.record_conventionality()

 def run_series(self, nr_interactions):
  for i in range(nr_interactions):
   self.run_interaction()
\end{lstlisting}

The \texttt{interact} method of the \texttt{NGInteraction} class defines the script according to which each communicative interaction takes place. A randomly selected agent, the speaker, formulates an utterance to draw the attention of another randomly selected agent, the hearer, to a randomly selected object in the environment, the topic. If there exists no construction in the speaker's grammar that associates a name with the topic object, the speaker calls its \texttt{learn} method to invent such a construction. The hearer then calls its \texttt{comprehend} method to retrieve the topic object in the environment, and its \texttt{learn} method in case it could not understand. The agents achieve communicative success if the hearer could identify the topic object, and both agents will positively or negatively reward their constructions at the end of the interaction, based on its outcome.

\begin{lstlisting}[language=Python]
class NGInteraction():
 ...
 def interact(self):
  s = self.speaker
  h = self.hearer
  s.utterance = s.formulate(s.topic)
  if s.utterance is None:
   s.learn(fcg.generate_word_form(), s.topic)
  if h.comprehend(s.utterance) is None:
    h.learn(s.utterance, s.topic)
  else:
   s.communicated_successfully = True
   h.communicated_successfully = True
  for agent in self.interacting_agents:
   agent.reward()
\end{lstlisting}

The \texttt{comprehend}, \texttt{formulate} and \texttt{reward} methods, as implemented for the NGAgent, illustrate how high-level PyFCG functionality facilitates the implementation of language game experiments. Not only do \texttt{comprehend} and \texttt{formulate} return the highest-scored solution, they also yield all competing solutions as a second return value. Successfully used constructions can then be rewarded positively through calls to their \texttt{increase\_score} method and their competitors can be rewarded negatively through calls to their \texttt{decrease\_score} method. Constructions that reach a score of 0 can be deleted from an agent's grammar using the \texttt{delete\_cxn} method. 

\begin{lstlisting}[language=Python]
def reward(self):
 if self.communicated_successfully:
  self.applied_cxn.increase_score(0.1)
  for cxn in self.competitor_cxns:
   cxn.decrease_score(0.2)
   if cxn.get_score() <= 0.0:
    self.delete_cxn(cxn)
 else:
  if self.discourse_role == 'speaker':
   self.applied_cxn.decrease_score(0.2)
   if self.applied_cxn.get_score() <= 0.0:
    self.delete_cxn(self.applied_cxn)
\end{lstlisting}

An experiment can be run by first creating a new instance of the \texttt{NGExperiment} class and then calling its \texttt{run\_series} method, passing the desired number of interactions as an argument: 

\begin{lstlisting}[language=Python]
>>> ng = NGExperiment({'nr_of_agents': 10, 'nr_of_objects': 5})
>>> ng.run_series(1500)
1500/1500 [100%] - 33.7s (44.50/s)
\end{lstlisting}

The results of an experiment can be visualised using any plotting library from Python's extensive ecosystem, such as \texttt{matplotlib}, \texttt{seaborn} or \texttt{plotly}. We demonstrate here the use of \texttt{matplotlib} to visualise the experiment's dynamics through graphs, where the degree of communicative success, degree of conventionality and average number of constructions are plotted in function of the number of interactions that have taken place \citep[see e.g.][]{vaneecke2022language}. 

\begin{lstlisting}[language=Python]
import matplotlib.pyplot as plt
pp = make_plot_points(MONITORS)
fig, axes = plt.subplots(len(pp.keys()))
for i, key in enumerate(pp.keys()):
    ax = axes[i]
    ax.plot(list(range(len(pp[key]))), 
       pp[key], label=key)
    ax.grid()
    ax.legend()
>>> plt.show()
\end{lstlisting}

\begin{figure}
\includegraphics[width=\columnwidth]{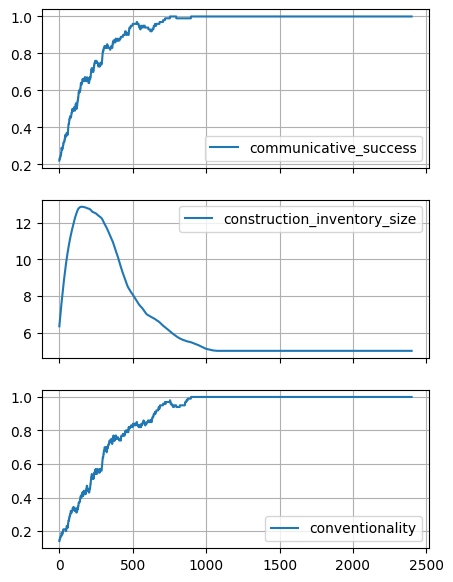}
\caption{Graphs showing the dynamics of a single run of the PyFCG-powered naming game experiment, featuring 10 agents and 5 objects, created using the matplotlib library.}
\label{fig:graph}
\end{figure}

Running this code yields the graphs presented in Figure \ref{fig:graph}, which show that the population indeed converges on a naming convention with one construction for each object in the world.

\section{Technical Implementation}

The design and technical implementation of PyFCG has been steered by two main considerations. First and foremost, the library needed to feel `native' to Python users, rather than familiar to users already accustomed to the reference FCG implementation. Second, the library needed to wrap the reference implementation, rather than reimplement the original codebase.

A crucial aspect contributing to the `native' look-and-feel of a Python library revolves around the library's embedding in the Python ecosystem of development tools. PyFCG is implemented as a Python package that is distributed via the Python Package Index (PyPI), and is hence `\textit{pip-installable}' on (at least) macOS, Linux and Windows. PyFCG's codebase is versioned using Git and distributed under an open source license, with its documentation being available via the \textit{Read the Docs} platform. Once installed, PyFCG exposes a range of high-level functions, classes and methods that build on common Python data structures such as objects, dictionaries, lists and tuples, thereby providing a truly `\textit{Pythonic}' interface to FCG and ensuring maximal compatibility with other Python libraries.

The choice to wrap the reference implementation rather than providing a reimplementation of FCG was motivated by two main reasons. First of all, by wrapping the original Common Lisp implementation, PyFCG can leverage its optimised, multi-threaded codebase and achieve comparable efficiency when running FCG's most compute-intensive procedures. The second reason was to avoid distributing development and maintenance efforts over two distinct FCG implementations, which could easily become counterproductive in the longer term. 

Technically, PyFCG provides a bridge to a stand-alone, executable version of FCG, named \textit{FCG Go}\footnote{https://gitlab.ai.vub.ac.be/ehai/fcg-go}. Upon calling PyFCG's \texttt{init} function, FCG Go is downloaded for the correct platform (if necessary) and launched as a background process. PyFCG communicates with this background process through HTTP requests, and thereby has access to the complete API of the reference implementation, as well as to its compiled codebase. When instances of PyFCG's \texttt{Grammar} and \texttt{Construction} classes are created, corresponding objects are instantiated in the FCG Go subprocess. Calls to methods that modify objects of these classes, such as \texttt{Agent.add\_cxn} and \texttt{Construction.set\_score}, modify their Python representation at the PyFCG side as well as their Common Lisp representation at the FCG Go side. Calls to functions and methods that more generally rely on functionality implemented by the FCG reference implementation, such as \texttt{fcg.start\_web\_interface} and \texttt{Agent.comprehend} are essentially rerouted to the running subprocess. Finally, functions and methods that do not modify FCG objects and are not compute-intensive, including those for listing an agent's constructions, for retrieving the scores of constructions, and for inspecting their internal representations, only involve objects on the Python side. 

Crucially, the architecture of PyFCG ensures that its users do not notice that the library interfaces with a non-Python subprocess. The subprocess is automatically launched when the library is initiated, without the need to install a Common Lisp environment or any other dependencies. PyFCG's  public interface is defined in terms of Pythonic data structures, any errors that would arise in the Common Lisp subprocess are caught and transposed to Python \texttt{Exception}s (subclass \texttt{FcgError}), and the subprocess itself is safely closed by a clean-up method when Python exits.

\section{Conclusion}

This paper has presented PyFCG as an open source software library that ports Fluid Construction Grammar to the Python programming language. PyFCG is publicly distributed as a pip-installable Python package, and provides a fully Pythonic interface to FCG's reference implementation. The library thereby enables its users to seamlessly integrate constructional language processing and learning into Python programs, and to combine this functionality with that of other libraries within Python's extensive ecosystem. At the same time, the design of PyFCG as a wrapper around an executable version of FCG's Common Lisp reference implementation, which stealthily runs its multi-threaded and compute-intensive procedures in the background, ensures that PyFCG can incorporate all FCG functionality and that the community's development and maintenance efforts can remain united.

Apart from a general description of the library, its motivation and its technical implementation, this paper has presented three walkthrough tutorials that showcase how PyFCG can be integrated in typical use cases of FCG. The first tutorial has demonstrated how FCG agents can be created, how they can be equipped with a human-designed grammar, how they can be instructed to comprehend and formulate natural language utterances, and how these processes can be visually inspected using FCG's web interface. The second tutorial has showcased how FCG agents can learn grammars from semantically annotated corpora, how they can use these grammars to annotate new data, and how grammars consisting of tens of thousands of constructions can efficiently be saved and later reloaded into agents. The final tutorial has demonstrated the use of PyFCG in setting up agent-based experiments on emergent communication. We have taken the example of the canonical naming game and shown how high-level FCG functionality can ease the implementation of the language processing and learning capacities of individual agents.

As we reach the end of this paper, it is useful to remind ourselves that Fluid Construction Grammar was conceived as an open instrument that provides a collection of high-level building blocks for constructional language processing, with the goal of supporting research and applications at the intersection of construction grammar and computational linguistics. The walkthrough tutorials described in this paper, along with their accompanying interactive notebooks, demonstrate example usage of the library, but are not meant to serve as guidelines or rule book in any way. Fluid Construction Grammar has always evolved to fit the needs of its users, and new perspectives or use cases brought by PyFCG users or contributors will be heartily embraced.

\section*{Acknowledgements}
We would like to thank our system administrator Frederik Himpe for his continuous, cross-platform technical and moral support, Arno Temmerman for his help in releasing PyFCG through PyPI and \textit{Read the Docs}, and Liesbet De Vos for designing the beautiful FCG Go logo.


\end{document}